\documentclass[acmtog,screen,authorversion]{acmart}

\usepackage{booktabs} 
\usepackage{subcaption,graphicx}

\usepackage{hyperref}
\usepackage{comment}
\usepackage{booktabs}
\usepackage{multirow}
\usepackage{mathtools}
\usepackage{cancel}

\usepackage{float}  
\usepackage{stfloats}  

\usepackage{color}
\definecolor{blue}{rgb}{0,0,0.6}
\definecolor{green}{rgb}{0,0.6,0}
\definecolor{red}{rgb}{0.6,0,0}
\definecolor{gray}{rgb}{0.4,0.4,0.4}
\definecolor{black}{rgb}{0,0,0}
\definecolor{lightgray}{rgb}{0.83, 0.83, 0.83}
\definecolor{purple}{rgb}{1,0,1}

\AtBeginDocument{%
  \providecommand\BibTeX{{%
    \normalfont B\kern-0.5em{\scshape i\kern-0.25em b}\kern-0.8em\TeX}}}

\setcopyright{acmlicensed}\acmJournal{TOG}
\acmYear{2022}\acmVolume{41}\acmNumber{4}\acmArticle{1}\acmMonth{7} \acmDOI{10.1145/3528223.3530179}

\acmBooktitle{ACM Trans. on Graph.}
\acmPrice{15.00}
\acmISBN{978-1-4503-XXXX-X/18/06}

\setcitestyle{square}

\begin{document}

\title[NeuralTailor]{NeuralTailor: 
Reconstructing Sewing Pattern Structures from 3D~Point Clouds of Garments}  

\author{Maria Korosteleva}
\affiliation{
    \institution{KAIST}
    \country{South Korea}}
\email{mariako@kaist.ac.kr}

\author{Sung-Hee Lee}
\affiliation{%
  \institution{KAIST}
  \country{South Korea}}
\email{sunghee.lee@kaist.ac.kr}

\renewcommand{\shortauthors}{Korosteleva and Lee}

\begin{abstract}


The fields of SocialVR, performance capture, and virtual try-on are often faced with a need to faithfully reproduce real garments in the virtual world. One critical task is the disentanglement of the intrinsic garment shape from deformations due to fabric properties, physical forces, and contact with the body. We propose to use a garment sewing pattern, a realistic and compact garment descriptor, to facilitate the intrinsic garment shape estimation.
Another major challenge is a high diversity of shapes and designs in the domain. The most common approach for Deep Learning on 3D garments is to build specialized models for individual garments or garment types. We argue that building a unified model for various garment designs has the benefit of generalization to novel garment types, hence covering a larger design domain than individual models would. We introduce NeuralTailor, a novel architecture based on point-level attention for set regression with variable cardinality, and apply it to the task of reconstructing 2D garment sewing patterns from the 3D point cloud garment models. Our experiments show that NeuralTailor successfully reconstructs sewing patterns and generalizes to garment types with pattern topologies unseen during training.

\end{abstract}





\begin{teaserfigure}
  \centering
  \includegraphics[width=0.97\linewidth]{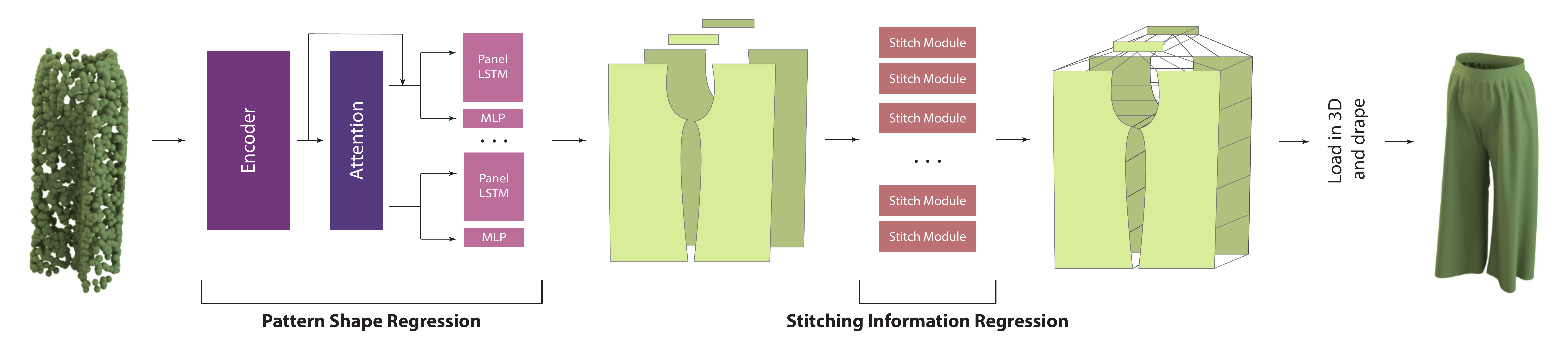}
  \caption{We introduce NeuralTailor -- a deep neural network-based framework that recovers structured representation of garment sewing patterns and stitching information from point clouds.}
\end{teaserfigure}
\maketitle
\section{Introduction}
\label{sec:intro}


Computer graphics has a long history of 3D garments research, from modeling to the physics simulation. In this paper, we tackle the problem of estimating the underlying rest shape of a garment for which a deformed 3D shape is available. Such a 3D garment shape could be a draping result from a physics simulation or a 3D scan of a real-world garment. 
By rest shape, we mean a representation of a garment shape disentangled from deformation due to external physical forces, collisions, and fabric properties.
Understanding such garment structure allows shaping the same garment in novel conditions like draping it on new body shapes or poses or enables the ability to adjust a captured garment's design. These abilities are highly desirable for virtual try-on, garment design, or avatar creation.

Building upon advances in Deep Learning research for shape analysis, we use a learning-based approach to tackle the problem in this work. To the best of our knowledge, ours is the first work to explore a learning-based solution for estimating garment rest shapes, with previous work either relying on fitting one of the pre-defined templates~\cite{Hasler2007ReverseGarments,Jeong2015c,Yang2018c}, optimizing rest shape directly from a good initial guess~\cite{Brouet2012b,Bartle2016c,Wang2018c}, surface flattening~\cite{Wang2009InteractiveCurves,Meng2012,Liu2018c,Bang2021EstimatingData}, or inverting physics deformations analytically~\cite{Ly2018} (see Sec.~\ref{sec:realted_work} for more detailed review). Learning-based methods have the advantage of scalability to acquire knowledge from large garment databases and cover the space of garment designs better than methods based on template fitting while allowing fast processing at inference time, unlike optimization-based solutions. 

One of our key ideas is to use a garment sewing pattern as a base representation for a garment rest shape. We assume a \textit{sewing pattern} to be a collection of the 2D pieces of fabric (panels) with a known placement of each panel around the human body and information on how the panels are stitched together to form the final garment. We model a \textit{panel} to be a closed piece-wise curve with every piece (\textit{edge}) being either a straight line or a Bezier spline.
Such a sewing pattern is a close approximation of how most real-world garments are constructed and thus serves as a strong prior for disentangling rest shape from physical deformation or the imperfections of the data acquisition process. At the same time, sewing patterns allow describing a variety of garment types and designs uniformly, unlike the approaches based on parametric templates as in~\cite{Wang2018}. Our problem formulation can also be viewed as a case of learning-based structure recovery for deformable objects. To the best of our knowledge, our work is one of the first of this kind, as most studies in the structure learning domain use datasets of rigid objects.

Reconstruction of a sewing pattern as a structure with a deep neural network (DNN) presents multiple challenges. It requires predicting a set (of panels) with variable cardinality across garment types. Every set element is a structured object itself and may exhibit significant shape variations, and stitches being cross-connections between individual edges of panels complicate the structure even more. We propose a NeuralTailor framework that recovers panel structure through a combination of 3D point-wise attention for a high-level decision on topology and an RNN module for predicting panel details. Stitching information is regressed using a separate stitch connection module implemented as a classifier on edge pairs. We also present a strong baseline model based on the RNN hierarchy to highlight the key features that enable NeuralTailor generalization properties. Our experiments show that NeuralTailor successfully reconstructs sewing patterns for various garment designs, including novel garments with sewing pattern structures not seen during training.

To summarise, our contributions are as follows: 
\begin{itemize}
    \item A new problem of learning-based recovery of a structured representation of garment sewing pattern.
    \item A strong baseline and an upgraded solution for the first deep learning framework that predicts a structural representation of a sewing pattern from a 3D garment shape and generalizes to novel garment types.
\end{itemize}

Code and pre-trained models for NeuralTailor framework are available at \href{https://github.com/maria-korosteleva/Garment-Pattern-Estimation}{GitHub}\footnote{ \url{https://github.com/maria-korosteleva/Garment-Pattern-Estimation}}.

\section{Related Work}
\label{sec:realted_work}

This section reviews previous studies related to our work. Specifically, we discuss research on estimating the rest shape of the garment, review other approaches to reconstructing controllable garments with DNNs, and approaches to representing structured objects in general.


\subsection{Garment Rest Shape Estimation} 
Several approaches have been proposed to estimate the rest shape of garments.

\textbf{Fitting garment templates.} The first approach is to rely on a set of garment templates and choose a template and its parameters to replicate an input garment as closely as possible. A notable work of~\cite{Chen2015c} performs a search in the database of 3D garment parts (skirts, sleeves, collars, etc.) and stitches them together to form the final garment. While having the advantage of consistently producing plausible garments, this approach requires an extensive database to deal with the high diversity of garments, which potentially leads to high computational costs. Several studies~\cite{Hasler2007ReverseGarments,Jeong2015c,Yang2018c} utilize sets of parametric sewing patterns, described in terms of numerical values such as sleeve length, waist width, etc., and perform optimization in the parameter space so that a chosen pattern matches an input garment when simulated on top of a body model. Wang et al.~\shortcite{Wang2018} takes one step further and trains a DL-model for each parametric garment template (defined similarly as in the studies mentioned above) to predict the template-specific parameters from an input garment sketch. Parametric templates reduce the storage demand and allow smooth exploration of the design space, unlike the databases of individual examples. However, different garment types would require different sets of parameters, forcing the usage of different models for each garment type and preventing knowledge sharing across types. To solve this, we propose using a sewing pattern as shape parametrization, a typical structure for most garment types. Our approach allows for generalization beyond the garment types present in the training set, as demonstrated by our experiments in Sec.~\ref{sec:evaluation}.


\textbf{Surface flattening.} Another way to get a garment's intrinsic structure is directly cutting a 3D surface of an input garment into developable 3D parts and then flattening every piece into 2D panels with, for example, ARAP technique~\cite{Igarashi2005} or Variational Surface Cutting~\cite{Sharp2018VariationalCutting}. The flattening-based approach works well in a controlled environment where an input 3D garment is not heavily deformed, the full, uncorrupted geometry is available, and when the initial cutting strategy is provided by a user or other means. For its simplicity and speed, this approach is popular in solutions for garment design where a high degree of manual control is acceptable~\cite{Wang2003FeatureSketches,Wang2005DesignProducts,Decaudin2006c,Yunchu2007PrototypeDummy,Daanen2008Made-to-measureScans,Wang2009InteractiveCurves,Meng2012,Liu2018c}. 
More recent studies adopt this approach to allow automatic processing by employing cuts guided by heuristic considerations~\cite{Bang2021EstimatingData} or by trained Deep Learning model~\cite{Goto2021Data-drivenGeometries}. These works produce plausible sewing patterns for various garment types but lack pattern quality and rely on the quality of original geometry a lot. Another recent work~\cite{Wolff20213DMovementc} relies on having direct access to the garment rest shape, which allows producing working patterns for arbitrary garment designs with a general technique of Variational Surface Cutting.
On the other hand, our learning-based approach neither requires uncorrupted, unsimulated geometry as input nor additional pre-processing and is capable of producing clean panel shapes.

\textbf{Pattern geometry optimization.} Much success in estimating garment structure in terms of sensitivity to details was achieved by optimization-based methods where some base geometric representation is deformed to achieve a target. The formulation of the target could be quite flexible, from some desired design features to a 3D garment model. Brouet et al.~\shortcite{Brouet2012b} demonstrated this approach to adapting garments to different body shapes while preserving the overall style. 
Bartle et al.~\shortcite{Bartle2016c} proposed a garment editing pipeline for users to create new garments directly in 3D by editing and combining existing garments while ensuring correct sewing patterns. Wang~\shortcite{Wang2018c} developed a method to adjust a standard sewing pattern for a better fit on an input body shape, while Li et al.~\shortcite{Li2018FoldSketch:Folds} enabled the creation of garments with desired folds' design by simple sketching over the initial model. More recently, Montes et al.~\shortcite{Montes2020ComputationalClothing} used optimization of sewing pattern geometry to find optimal fit and pressure distribution for tight clothing.
These methods require a good initial guess of the garment sewing pattern, which is often unavailable. Our approach follows a more loose assumption that the input belongs to a distribution modeled by the training data and even makes successful predictions on garment types outside the training domain. This assumption will become even less demanding as more data becomes available.

\textbf{Inverting Physics}.  The work of \cite{Ly2018} explores an interesting direction of performing a physics inversion by jointly estimating the rest shape and the physical forces acting on an input object conditioned on material properties provided by the user. This approach applies to any shell-like objects, including garments, and is not limited by the representational power of datasets. On the other hand, the proposed method is computationally demanding and has trouble handling the folds and wrinkles due to contacts, which is typical for garments draped on humans. On the contrary, our approach successfully  processes folds and wrinkles
and can perform fast once trained. 



%
\subsection{Learning-based reconstruction of controllable garments}

We see a potential for our approach of predicting sewing patterns to be used for reconstructing controllable 3D garments from 3D scans or images of people. A number of works in recent years address this problem for learning-based retargeting~\cite{Wang2018,LalBhatnagar2019,Santesteban2019,Bertiche2020,Ma2020,Patel2020,Zakharkin2021Point-BasedClothing},  animation~\cite{Wang2019,Patel2020,Jiang2020,Ma2021TheClothing,Zakharkin2021Point-BasedClothing,Santesteban2021Self-SupervisedTry-On}, or garment style adjustment~\cite{Wang2018,Tiwari2020,Su2020DeepClothEditing,Corona_2021_CVPR}. Some of the works rely on meshes of known topologies and thus require their models to be trained per-garment or per-garment type~\cite{Wang2018,Wang2019,Santesteban2019,Jiang2020,Tiwari2020,Patel2020}. 
Usage of displacements~\cite{LalBhatnagar2019,Bertiche2020,Ma2020}, UV-maps~\cite{Su2020DeepClothEditing}, point clouds~\cite{Ma2021TheClothing,Zakharkin2021Point-BasedClothing}, and implicit functions~\cite{Corona_2021_CVPR} enabled representation of different garment styles within the same model and even showed the ability to reconstruct unseen outfits~\cite{Ma2021TheClothing}, but the garments are reconstructed with design, material properties and deformations fused together. 

We believe that reconstructing disentangled garment representations will eventually lead to better quality, control, and generalization. Shen et al.~\shortcite{Shen2020GAN-BasedImages} demonstrate a garment model generator conditioned on sewing patterns with capabilities to generalize to novel designs. In our work, we show that using sewing pattern as a natural structured representation of design when inferring it from raw inputs allows not only for generalization to \emph{unseen garment examples} as in~\cite{Ma2021TheClothing}, but \emph{unseen garment types}. Moreover, sewing patterns are retargetable by design and, when coupled with a physics simulator, produce physically accurate 3D reconstructions with guaranteed developability, which geometry-based learned reconstructions cannot do yet.  

\subsection{Structural Deep Learning} 

The problem of representing sewing patterns in DNN is highly related to a more general problem of representing the structure of the objects composed of simpler components in DNN. Studies on this problem often experiment with 3D furniture as an example of such structured objects. Our work builds upon the ideas of hierarchical and sequential modeling of part relationships of GRASS~\cite{Chaudhuri2017}, StructureNet~\cite{Mo2019StructureNet:Generation}, SAGNet~\cite{Wu2019SagNet:Modeling}, LSD-StructureNet~\cite{Roberts2021LSD-StructureNet:Hierarchies}, and several works in vector graphics generation~\cite{Ha2018ADrawings,Carlier2020DeepSVG:Animation,Wang2021DeepVecFont:Learning}. On top of that, we attempt to generalize beyond the component collections presented in the training set. Hence, we introduce a novel attention module and connectivity classification module for stitch prediction. A recent work of Shape Part Slot Machine~\cite{Wang2021TheParts} demonstrated a similar generalization ability by focusing on connections between components rather than the global shape. In our work, we come to a similar conclusion as \cite{Wang2021TheParts} -- that focusing on the local context allows for both prediction quality and structural generalization capabilities.

\section{Dataset}
\label{sec:dataset}

In this work we use Dataset of 3D garments with sewing patterns  ~\cite{garment_pattern_dataset} as introduced in~\cite{Korosteleva2021GeneratingPatterns}. It covers a variety of garment designs, including variations of t-shirts, jackets, pants, skirts, jumpsuits, and dresses, with 22,000 garments sampled from 19 base types in total. Each garment sample contains a garment 3D model as draped on SMPL~\cite{Loper2015} average woman body shape in T-pose, a corresponding sewing pattern represented as a structure, and a corrupted 3D model imitating some of the 3D scanning artifacts. The dataset is limited in representing human poses and shapes but provides a good range of garment designs. Hence, it provides a good starting point for tackling the problem of sewing pattern recovery from 3D models.

\subsubsection*{Panel classes and panel vectors}
\label{sec:data:classes}

The original dataset does not guarantee that similar panels from different garment types, e.g., pant panels in pants and jumpsuits, have the same labels. Hence, we introduce classes of panels that we use to group panels by role and location around the body across garment types. For example, panels covering the front of the trunk from T-Shirts, dresses, jumpsuits, etc., are grouped in the "front panels" class. The labeling for panels of base templates is included with the published code.

\subsubsection*{Additional sample filtering}
\label{sec:data:filter}

\begin{figure}[t]
  \centering
  \includegraphics[width=0.8\linewidth]{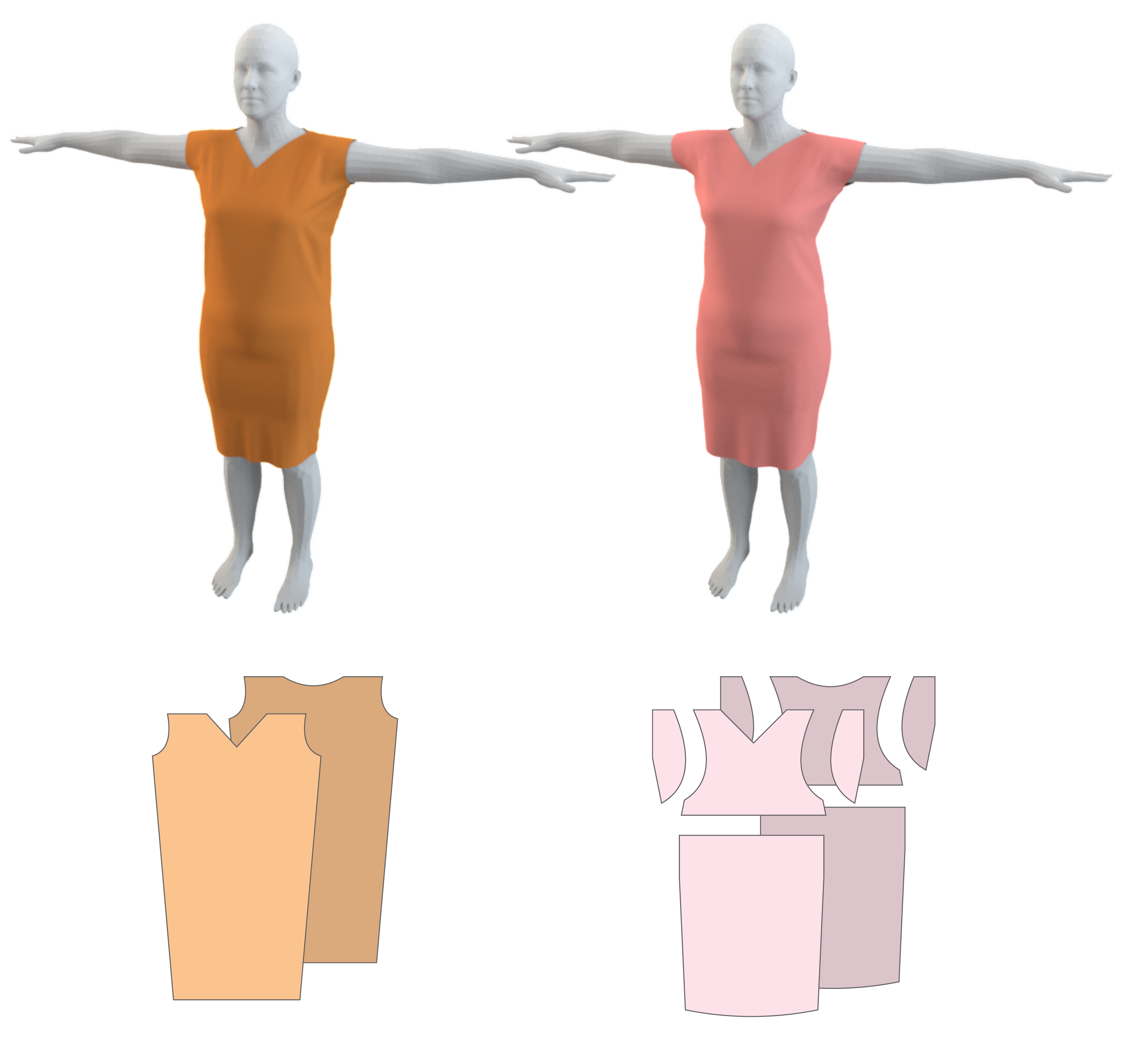}
  \caption{Garments with design overlap. Two different sewing patterns produce similar 3D shapes when draped.}
  \label{fig:data:design_overlap}
\end{figure}

The original dataset contains garment samples with overlapping designs -- these samples have different sewing pattern topologies and may belong to different garment types but produce similar shapes in 3D, as shown in Fig.~\ref{fig:data:design_overlap}. Such cases are common in real-world garments, but they significantly complicate an already difficult problem. In this work, we assume that design overlap in the data is minimal to focus our attention on developing a base solution for sewing pattern reconstruction and topology generalization. We manually analyzed parameter spaces of the dataset base templates and filtered them to contain mostly non-overlapping examples.

\subsubsection*{Dataset split}
\label{sec:data:split}

We use the train/test group split of garment types as introduced in the dataset~\cite{garment_pattern_dataset}. Garment samples from the seven types of the test group remain unseen to the NeuralTailor during training and are only used for evaluation, as shown in Sec.~\ref{sec:evaluation}. We additionally designate 100 examples of each train group type as a validation set for model selection and 100 examples of each type as a test set to compare performance on seen and unseen types. This split leaves 19236 garment samples in the full training set. The number of training samples when sample filtering is applied is 9678.

\section{Overview}
\label{sec:overview}

Our work explores several directions to approach the task at hand. We designed an original thought-through baseline model that can successfully represent the sewing pattern structures and learn to reconstruct them (Sec.~\ref{sec:baseline_model}). It is based on extracting a latent space vector from an input point cloud and then decoding it into a sewing pattern through a two-step hierarchy of RNNs, with stitching information represented as a property of individual panel edges.

We then updated the baseline with several new ideas (Sec.~\ref{sec:NT_framework}). Firstly, we introduce a point-level attention mechanism that evaluates latent codes for individual panels based on local rather than global context. Secondly, we separate stitch prediction into an independent module that performs edge pairs classification into being connected or not.

As shown in Sec.~\ref{sec:evaluation}, these improvements make the overall framework generalize to sewing pattern topologies unseen during training. The achieved generalization feature is critical for a diverse domain like garments. Gathering a fully representative dataset of garments is genuinely hard, and it is nearly impossible to achieve design generalization by training specialized per-type models, which is a common approach in Deep Learning for 3D garments.

\section{Baseline Model}
\label{sec:baseline_model}


\begin{figure*}[ht]
  \centering
  \includegraphics[width=\linewidth]{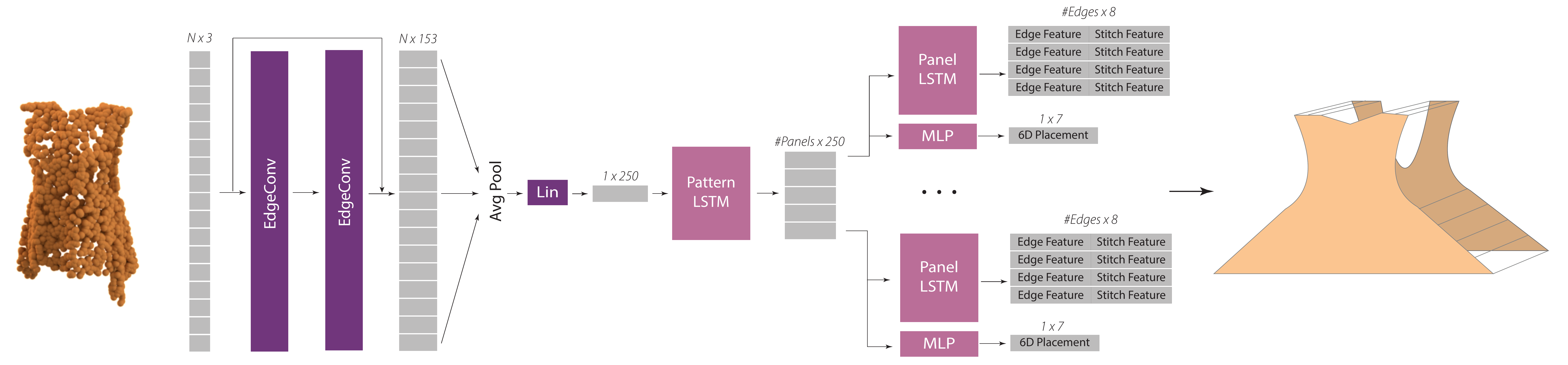}
  \caption{Overview of Baseline architecture for sewing pattern recovery. The input point cloud of $N$ points gets processed by two EdgceConv~\cite{Wang2018DynamicClouds} layers outputting per-point features that are then aggregated by average pooling to form the latent garment code. This latent representation is then decoded into the predicted sewing pattern by hierarchically organized LSTM blocks. Pattern-level LSTM unfolds the latent code into the list of encodings of individual panels. Each panel encoding is processed by another LSTM cell to produce a sequence of edge features (including stitching information) and 6D placement of the panel. The same Panel LSTMs and MLP models are applied to each panel encoding.}
  \label{fig:net:baseline}
\end{figure*}

\subsection{Point Features Encoder}
\label{sec:baseline:point_encoder}

Deep learning-based processing of point clouds is a challenging problem of its own and has seen rapid development in recent years. In this work, we employ EdgeConv~\cite{Wang2018DynamicClouds} as a base block for the encoder for its simplicity and performance on par with other state-of-the-art point cloud-based network architectures, according to~\cite{Guo2020DeepSurvey}. The main advantage of EdgeConv is its ability to aggregate information in feature space rather than spatially by dynamically re-building a connectivity graph on every EdgeConv layer. Our encoder consists of two EdgeConv layers, as shown in Fig.~\ref{fig:net:baseline}, 
with a skip-connection from the input 3D point cloud to the output of the last EdgeConv layer. Final per-point features are then aggregated into a single feature vector by average pooling.

\subsection{LSTM for Panel Encodings}
\label{sec:baseline:rnn_enc}

In the next stage, the model reconstructs latent codes for each panel in a sewing pattern given the global latent code. Having no particular order with respect to each other, panels represent a set whose cardinality (number of panels) varies across garments. One of the simpler solutions to represent such a structure is to define an arbitrary order on set elements and then employ a sequence-based model to predict the elements of the formed sequences, as was done for structured predictions in ~\cite{Wu2019SagNet:Modeling}. OrderlessRNN~\cite{OguzYazici2020OrderlessClassification} takes the set-as-sequence approach even further by allowing an RNN-based network to output set elements in any order instead of following a pre-defined one by a clever loss function construction, which helps improve the performance of their target task.

Following this line of work, we design panel encoding prediction as an LSTM~\cite{Hochreiter1997LongMemory} module for our baseline model for its capability to model sets of variable cardinality. This LSTM module takes in a garment global latent code and outputs a sequence of latent vectors of panels, which are then processed by Panel Decoder as described below. 
We experiment with both pre-defined order and orderless losses approach, as shown in Sec.~\ref{sec:evaluation}.

\subsection{Panel Decoder}
\label{sec:baseline:panel_dec}

Each panel latent vector predicted in the previous step is processed by the panel decoder to recover the panel shape 
and stitching information. The panel decoder consists of an LSTM module and a linear module. The LSTM converts a panel encoding into a sequence of edge features comprising the panel, and an additional linear module regresses the 3D placement of the panel. The details of edge and placement representations are discussed below.

\subsubsection{Panel representation.}
\label{sec:baseline:panel_def}

We model a panel as a sequence of edges -- smooth segments of a closed piece-wise curve -- with every edge being either a straight line or a quadratic Bezier spline, similarly to their representation in the dataset. Using splines to represent curvy edges instead of discretization is more compact, prevents resolution-related artifacts, and ensures simple stitch definition as a 1-to-1 edge connection.

\subsubsection{Edge features.}
\label{sec:baseline:edge_feature}

We use the following idea to construct a meaningful sequential representation of panel edges. Panel decoder outputs every edge as \textbf{a 2D vector}, from the edge starting point to its endpoint as follows:
\[
    \vec{e}_{ij} = v_j - v_i,
\]
where $v_i$ and $v_j$ are the 2D local coordinates of vertices $i$ and $j$ connected by $e_{ij}$. Since every panel is a closed piece-wise curve, these edge vectors form a loop when ordered and traced sequentially. 2D coordinates of any panel vertex can be obtained by adding a corresponding edge vector to the 2D coordinates of a previous vertex in the panel. The first vertex of the loop is always assumed to be at the origin of the panel local space. The dataset guarantees consistency in the choice of edge loop first vertex and the direction of loop traversal across panels by design, so we simply use the edge loop order as given in the data when evaluating losses.

Since edges are not necessarily straight lines, we use \textbf{curvature coordinates} as an additional edge vector feature. Curvature coordinates are the 2D coordinates $(c_x, c_y)$ of quadratic Bezier spline control point and are defined in the local space of an edge. In this coordinate system $(0, 0)$ and $(1, 0)$ indicate the positions of edge vertex. Hence, $c_x$ indicates the position along the edge, roughly corresponding to the location of the curvature peak, and $c_y$ controls the depth of the curvature. If an edge is straight, its curvature coordinates are marked as $(0, 0)$. The edge feature will then look like this:
\[
    (e_x, e_y, c_x, c_y),
\]
where $(e_x, e_y) = \vec{e}_{ij}$ are 2D edge vector coordinates and $(c_x, c_y)$ are curvature coordinates.

Since different panels have different numbers of edges, the edge sequences are padded with zero feature vectors to the length (14 in our experiment) that is equal to or larger than the maximum number of edges found in the training set.

\subsubsection{Stitch Tags for stitching information prediction}
\label{sec:baseline:stitch_tags}

Stitches are cross-connections of the edges in the network output hence predicting them represents a challenge for a feed-forward style of network architecture. Our first idea is to include stitching information directly into the edge features. We define per-edge stitching information as the following feature vector:
\[
    (f_{0/1}, s_1, s_2, s_3),
\]
where $f_{0/1}$ is a \textbf{binary class} of whether an edge is free or belongs to any of the stitches, and $(s_1, s_2, s_3)$ is a learned vector called a \textbf{stitch tag} that is designed to identify the edges that are connected. The definitive property of stitch tags is as follows:
tags of edges from the same stitch are expected to be similar, but edges from different stitches should have tags that are different by a margin. We use Euclidean distance between tags as a similarity measure. The connectivity reconstruction then comes down to filtering out free and connected edges and comparing stitch tags of the pairs of connected edges. Note that the edges classified as ``free'' are not expected to have meaningful stitch tags.

This idea enables a compact representation of pattern connectivity that does not depend on the number of stitches or the total number of edges in a pattern and avoids explicitly referencing edge IDs, allowing the encoding of different sewing pattern topologies. The network learns to provide suitable values of the stitch tags by following the loss function that enforces the correct behavior during training, as described in Sec.~\ref{sec:baseline:losses}.

\subsubsection{Panel 3D placement representation.}

The following feature vector represents the panel placement in the world space:
\[
   (q_1, q_2, q_3, q_4, t_1, t_2, t_3),  
\]
where $(q_1, q_2, q_3, q_4)$ is a \textbf{quaternion} that reflect panel rotation. Panel translation $(t_1, t_2, t_3)$ is represented as \textbf{3D translation} of the top mid-point of the panel's 2D bounding box when the panel is viewed in 3D. We found that in most cases, this point corresponds to body features (e.g., neck, waist) important for panel placement and thus exhibits stability across particular stylistic choices (e.g., skirt length). This translation formulation showed a more accurate 3D placement prediction than using panel local origin as the reference point in our tests.

\subsection{Loss functions}
\label{sec:baseline:losses}

The full loss for training the panel shape and placement prediction module is as follows:
\begin{equation}
\label{eq:full_loss}
    L_{total} = L_{edge} + L_{loop} + L_{placement} + L_{stitches}
\end{equation}

\subsubsection{Edge loss}
The edge loss $L_{edge}$ evaluates the quality of panel geometry prediction.
Ground truth panel representation is converted to the sequential format of 2D edge vectors as described above. Then, $L_{edge}$ is computed as an MSE loss on edge vectors and curvature coordinates between the ground truth and the corresponding edge features from NeuralTailor output.
\subsubsection{Loop loss}
The loop loss $L_{loop}$ is added to additionally enforce the loop closure property of panel representation. It evaluates the $L_2$ norm of the distance between the origin and final point of the panel edge sequence.

\subsubsection{Placement loss}
The placement loss $L_{placement}$ is an MSE loss on corresponding rotation and translation representations converted from ground truth placement information to match the network output specification.

\subsubsection{Losses for stitch prediction}
\label{sec:net:losses:tags}
Training loss for predicting stitching information consists of the following two terms:

\begin{equation}
\label{eq:stitch_loss}
    L_{stitches} = {L_{class} + L_{tags}}.
\end{equation}

The class loss $L_{class}$ encourages learning proper edge class, modeled as a binary cross-entropy loss for edges classification into free and non-free. The tag loss $L_{tags}$ enforces the definition of the stitch tags (Sec.~\ref{sec:baseline:stitch_tags}) by referencing a list of stitches from the data. Its formulation is a variation of the triplet loss~\cite{Schultz2003} and consists of two components -- similarity and separation losses. $L_{similarity}$ encourages the stitch tags of a pair of edges that are stitched together to be close to each other, and $L_{separation}$ pushes all the stitch tags from different stitches apart by a predefined margin $\delta$ as follows:
\begin{align*}
\label{eq:tag_loss}
    L_{tags} &= L_{similarity} + L_{separation} \\
    L_{similarity} &= \sum_{\mathclap{(i, j) \in stitches}}
                    \left \| tag_{i} - tag_{j} \right \|^2  \\
    L_{separation} &= \sum_{\mathclap{
                        \substack{i, j \in non\_free \\ 
                                (i, j) \notin stitches}}}
                    max(\delta - \left \| tag_{i} - tag_{j} \right \|^2, 0),
\end{align*}
where $i$, $j$ are edge IDs, $stitches$ is a set of edge pairs to be stitched together, and $non\_free$ is a set of non-free edges that participate in any of the stitches, as opposed to the edges that are left free. Both sets are obtained from the ground truth sewing pattern. 

We found the training to be more efficient if $L_{class}$ and  $L_{tags}$ losses are introduced after a few epochs, allowing the model to learn the overall concept of sewing patterns first. In our experiments, these loss components are added after the 40th epoch. 

\subsubsection{Implementation of panel ordering and padding}

Evaluation of the above losses requires a choice of panel ordering within a sewing pattern. To ensure that ground truth panels of the same class are matched to the same positions in the net output panel sequence, we organize panels within sewing patterns into panel vectors. We fix the order of panel classes and place each existing panel in the panel vector according to its class id. The slots corresponding to classes not present in a sewing pattern are filled with empty panel placeholders, 
represented by zero tensors of the same dimensionality as the actual panels. In contrast to the usual approach of placing the padding at the end of a sequence, this arrangement spreads the panel placeholders across the panel sequence. This strategy allows the ordering to be more consistent across different topologies and encourages the network to explore the similarity between the panels from the same class. In our experience, the choice of padding strategy did not seem to affect the performance of the Baseline model, but it paid off when we improved the architecture, as discussed in Sec.~\ref{sec:NT_framework}.

We additionally experiment with removing the panel order by finding the order of panels in the ground truth that best matches the order of panels in predicted sewing patterns, similarly to OrderlessRNN~\cite{OguzYazici2020OrderlessClassification}. The matching is performed by solving an assignment problem between the two sets of panels with an off-the-shelf algorithm. The distances between panels are evaluated as Euclidean distances of their vector representation consisting of the concatenation of all the edge features in a panel and 6D placement vector.
\section{NeuralTailor: Improvements for generalization}
\label{sec:NT_framework}


\begin{figure*}[ht]
  \centering
  \includegraphics[width=\linewidth]{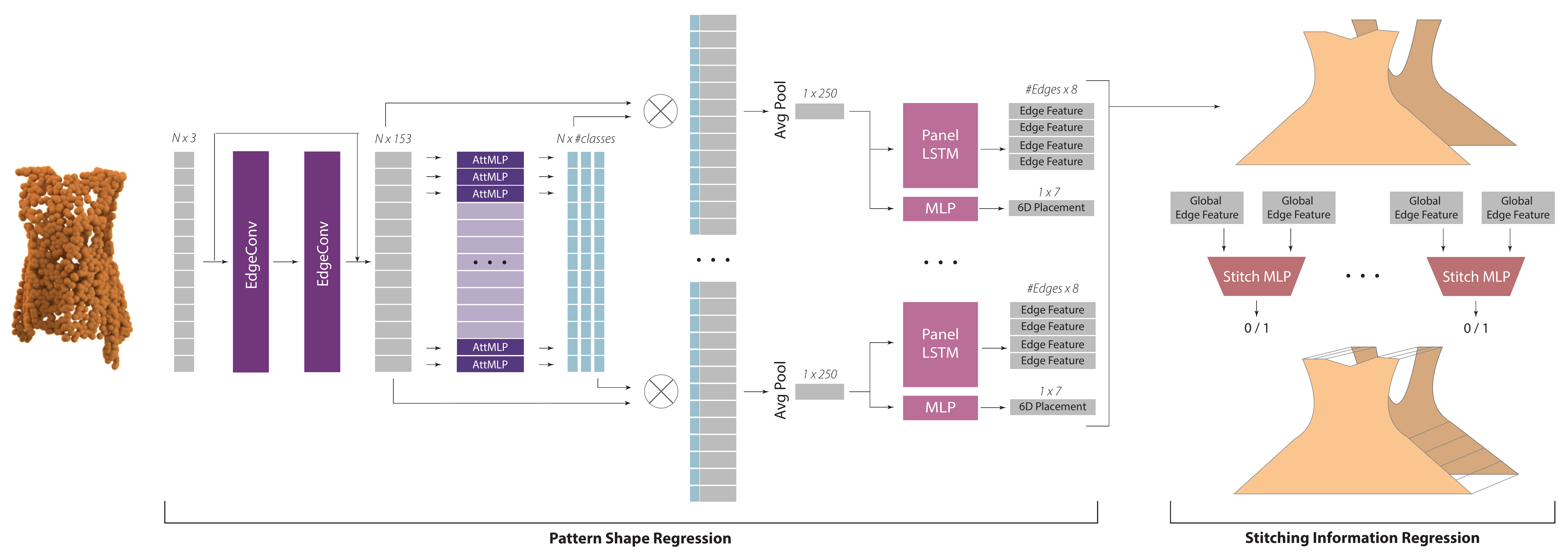}
  \caption{Overview of NeuralTailor architecture. The first core difference with baseline is the attention MLP module that predicts per-panel-class attention scores for each point (MLP weights are shared across the points). The scores are then used to weigh the point features before aggregating them into per-panel latent codes that are then decoded into panel shapes as in the baseline. The second difference is the recovery of stitching information by using a separate StitchMLP that classifies edge pairs of the predicted pattern into being stitched or not (MLP weights are shared across edge pairs). }
  \label{fig:net:neural_tailor_full}
\end{figure*}  

We introduce two modifications to the baseline framework that encourage modular and local context-based reasoning within the architecture. These features enable recognizing familiar panel components and reconstructing novel pattern topologies by recombining these components.

\subsection{Attention-based Panel Encodings}
\label{sec:net:attention}


\begin{figure*}[ht]
  \centering
  \includegraphics[width=\linewidth]{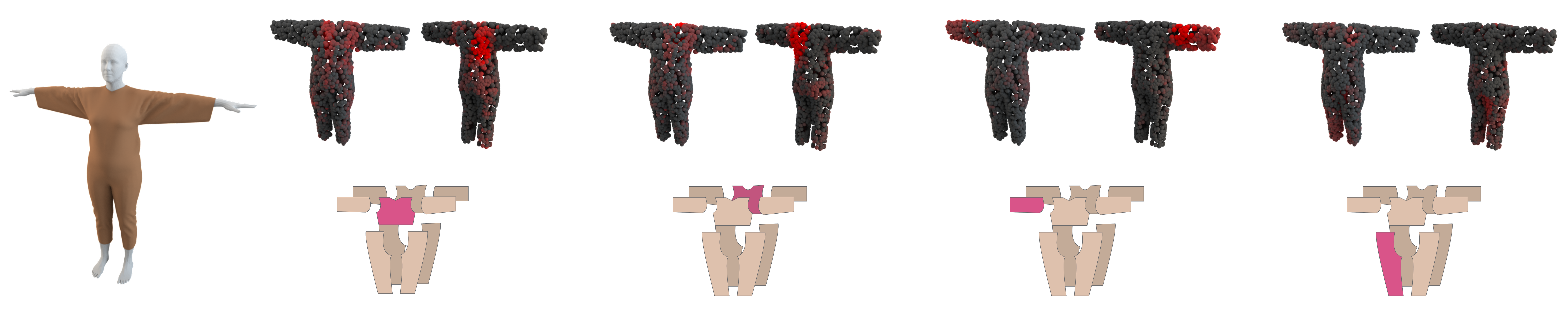}
  \caption{Attention weights for a jumpsuit example (unseen type) corresponding to highlighted panels. Interestingly, attention weights spread to the areas of panels' symmetrical counterparts (sleeve, top) and, when needed, to the representative areas of global shape (shape of the top front and back panels depends on whether the bottom is skirt or pants, hence the model grabs some features from the bottom. Best viewed on screen zoomed-in.)}
  \label{fig:att_weights}
\end{figure*}  

As introduced in Sec.~\ref{sec:baseline_model}, the baseline model employs sequence prediction from global garment latent code to reconstruct the latent codes for individual panels. This global bottleneck makes the model prone to relying on the overall shape of the garment and less likely to exploit its per-component structure. 

Another option would be to attend to only a relevant part of the input point cloud to construct a latent code for a corresponding panel. This approach would allow the model to construct the final sewing pattern from relevant pieces. An additional consideration is that 
garments of different types have a different number of panels, and the types of panels differ, too.

One thing we may safely assume is that one sewing pattern contains no more than one panel of each panel class as per the definition of panel classes (Sec.~\ref{sec:data:classes}). Hence, we implement the attention idea by introducing an additional MLP module that acts on point feature vectors and predicts per-point per-panel-class probability scores of how likely it is that a particular point belongs to a given panel class, as shown in Fig.~\ref{fig:net:neural_tailor_full}. The model then obtains encodings for each panel class by simply pooling point features as given by the encoder (Sec.~\ref{sec:baseline:point_encoder}) weighted by the attention scores of the relevant class. 

Ideally, attention weights should encourage the recognition of components corresponding to panel classes across different garments. Hence, we would like attention weights to highlight only the minimal local context relevant to each panel class. For that reason, weights should be sparse, and each point should participate in just one or a few classes. We encourage this behavior by employing SparseMax~\cite{pmlr-v48-martins16} as the last layer of the attention module evaluated on per-point attention scores. 

Examples of attention weights predicted by our framework are given in Fig.~\ref{fig:att_weights}. The attention weights highlight the local context and contain additional clues related to garment type.

\subsection{Stitching Information Regression Neural Network}
\label{sec:net:stitch_model}

Stitch-tag-based pattern stitches prediction of the baseline model yields an effective representation of stitches to reconstruct the whole sewing pattern in one single model. 
However, both stitching information and panel shape are inferred from the same panel latent code, which may give an entangling effect of the two properties. For example, front panels of T-shirts and jumpsuits have only one difference -- the latter requires the bottom edge to be separated into two for correct connection with pant panels. Although the training set contains examples of T-shirt front panels connected to sleeves, it does not have such examples for jumpsuits. Having stitching information entangled with shape may result in the network replicating this bias of the training data and refusing to predict stitches for jumpsuits front panels even when sleeves are present in the input 3D point cloud.

For this reason, we shifted towards considering the panel edges as individual objects rather than panels' parts. We hypothesize that the pattern geometry and panel placement may provide enough information to predict the stitches without accessing the input geometry. Surprisingly, it turned out to be true.  

We constructed a simple MLP model that takes a pair of sewing pattern edges as input and outputs the probability of these edges being connected by a stitch (Fig.~\ref{fig:net:neural_tailor_full}). Each edge is represented as a vector
\[
    (v_x^{start}, v_y^{start}, v_z^{start}, v_x^{end}, v_y^{end}, v_z^{end}, c_x, c_y),
\]
where $(v_x^{start}, v_y^{start}, v_z^{start})$ and $(v_x^{end}, v_y^{end}, v_z^{end})$ are 3D coordinates relative to the body model of panel vertices connected by the stitch, and $(c_x, c_y)$ are the edge curvature control point coordinates, as described in~\ref{sec:baseline:edge_feature}. 

\subsubsection{Training set structure}
\label{sec:stitch_model:training set}

The only tricky part of training this stitch classification model was setting up the training set. A naive training set would include all possible combinations of edge pairs for each sewing pattern in the garment dataset. This training set is highly unbalanced as most edge combinations are not connected by a stitch. It also has inadequately more examples for complex patterns with many panels than simpler ones, as the number of edge pairs grows quadratically with the total number of edges in a sewing pattern. The latter property also results in the fast growth of the training set size as we add more sewing patterns to it. Instead, on each epoch, we sample a given number of edge pairs from each sewing pattern, with oversampling of stitched pairs and under-sampling of non-connected pairs. Since our dataset contains many samples that share sewing pattern topologies, we expect the network to get enough clues for non-connected pairs during training. In addition to these precautions, we avoid bias towards a particular choice of vertices or edges order in pairs by randomizing these properties at training time.

\subsection{Training process adjustments}

Training of NeuralTailor framework is now performed in two steps: first, training the model for pattern shape regression, and second, training the stitch prediction model, in this order. The stitch prediction model is trained on the edge features reconstructed by the pattern shape model instead of using edges from ground truth sewing patterns. This decision increases robustness to noise in the pattern shape model output at inference time, which we demonstrate in Sec.~\ref{sec:eval:stitch_train_set}. 

The losses for each model are adjusted as follows. Since the stitch prediction moves to a separate module, the $L_{stitches}$ is not needed for training the pattern shape regression; hence the total loss for it is:
\begin{equation}
\label{eq:shape_loss}
    L_{pattern shape} = L_{edge} + L_{loop} + L_{placement},
\end{equation}
with $L_{edge}$, $L_{loop}$, and $L_{placement}$ being the same as introduced in~Sec.~\ref{sec:baseline:losses}. The ordering of ground truth panels for loss evaluations follows the same scheme based on panel classes as for the baseline model training. As for stitch regression model, we train it as a binary classification task using binary cross-entropy loss (BCE).

\section{Evaluation}
\label{sec:evaluation}

\begin{table*}[ht]
\centering
\begin{tabular}{@{}lcccccccccc@{}}

\toprule
 &
  \multicolumn{5}{c}{Seen types} &
  \multicolumn{5}{c}{Unseen Types} \\
 &
  Panel L2 &
  \#Panels &
  \#Edges &
  Rot L2 &
  \multicolumn{1}{c}{Transl L2} &
  Panel L2 &
  \#Panels &
  \#Edges &
  Rot L2 &
  Transl L2 \\ \midrule
  
LSTM w Stitch Tags &
  5.35 & 
  98.3\% & 
  99.0\% & 
  0.02 & 
  \multicolumn{1}{c|}{1.04} & 
   20.4 & 
   1.6\% & 
   35.3\% & 
   0.25 & 
   9.02 \\  
   
LSTM &
  2.71 &
  \textbf{99.8\%} &
  \textbf{99.9\%} &
  \textbf{0.004} &
  \multicolumn{1}{c|}{\textbf{0.32}} &
  14.7 &
  6.5\% &
  53.2\% &
  0.17 &
  6.75  \\
  
LSTM Orderless  &
  2.87 &
  99.4\% &
  \textbf{99.9\%} &
  \textbf{0.004} &
  \multicolumn{1}{c|}{0.33} &
  12.94 &
  2.7\% &
  59.0\% &
  0.16 &
  7.18 \\

Att (NeuralTailor) &
  \textbf{1.5} &  
  99.7\% &  
  99.7\% &  
  0.04 &  
  \multicolumn{1}{c|}{1.46} &  
  \textbf{5.2} &  
  83.6\% & 
  87.3\% &   
  \textbf{0.07} &    
  3.22 \\ \hline  
  
Att on Scan Imitation &
   1.53 &   
   99.6\% &
   99.8\% &
   0.04 &
  \multicolumn{1}{c|}{1.42} &
   5.44 &
   83.2\% &
   86.3\% &
   0.07 &
   3.39 \\

Att w Stitch Tags &
  1.83 &
  99.6\% &
  99.8\% &
  0.05 &
  \multicolumn{1}{c|}{1.67} &
  8.31 &
  69.9\% &
  79.9\% &
  0.08 &
  3.42 \\
  
Att w/o Loop Loss &
  1.62 &
  98.7\% &
  87.5\% &
  0.04 &
  \multicolumn{1}{c|}{1.53} &
  5.97 &
  \textbf{85.5\%} &
  60.9\% &
  0.07 &
  3.35 \\
  
Att w Segm &
  2.28 &
  95.3\% &
  99.6\% &
  0.05 &
  \multicolumn{1}{c|}{1.62} &
  7.48 &
  76.3\% &
  79.5\% &
  0.07 &
  \textbf{2.87} \\ 

Att w Alt Classes &
  1.53 &
  98.8\% &
  99.6\% &
  0.04 &
  \multicolumn{1}{c|}{1.45} &
  7.96 &
  73.1\% &
  80.5\% &
  0.08 &
  3.57  \\

Att w/o Data Filter &
  1.6/1.95 &
  98.6/97.5\% &
  99.8/99.2\% &
  0.07/0.07 &
  \multicolumn{1}{c|}{2.2/2.5} &
  6.2/6.4  &
  81.6/75.2\% &
  \textbf{88.5}/88.2\% &
  0.08/0.10  &
  3.9/4.5 \\

  \bottomrule
\end{tabular}
\vspace{5pt}
\caption{Evaluating quality of pattern shape prediction on a test set consisting of garment types seen and unseen during training in various experiments. The results for \emph{Att w/o Data Filter} are reported for filtered and non-filtered test sets divided by "/".}
\label{tab:shape_eval}
\end{table*}

\begin{table}[t]
\centering
\begin{tabular}{@{}lcccc@{}}
\toprule
 & \multicolumn{2}{c}{Seen types} & \multicolumn{2}{c}{Unseen Types*} \\ \midrule
 & Precision & Recall & Precision & Recall \\ \midrule

Stitch Tags & 
    \textbf{99.9\%} & 
    \textbf{99.9\%} & 
    70.1\% & 
    72.9\% \\
    
Model on GT& 
    96.6\% & 
    88.6\% & 
    \textbf{75.3\%} & 
    60.6\% \\
    
Model on Predictions &
    96.3\% & 
    \underline{99.4\%} & 
    74.7\% & 
    \textbf{83.9\%} \\ \bottomrule
\end{tabular}
\vspace{5pt}
\caption{Evaluation of stitch prediction in different experiments. Evaluation of the Stitch Model is performed on sewing patterns reconstructed from 3D inputs by the pattern shape model. * To reduce error propagation, performance for unseen garment types was evaluated only on sewing patterns with a correctly predicted number of panels.}
\label{tab:stitch-eval}
\end{table}


This section demonstrates the capabilities of NeuralTailor in different setups. First, we introduce the collection of measurements to evaluate sewing pattern prediction quality. We then compare the performance of NeuralTailor with our baseline solutions for pattern shape and stitch information reconstruction on garment types that were used or hidden during training. We then further analyze the behavior of the framework with different loss conditions, the effects of changing the panel classes or removing dataset filtering (as introduced in Sec.~\ref{sec:dataset}), as well as the potential for generalization to in-the-wild data. Lastly, we compare sewing patterns predicted by NeuralTailor with the patterns suggested by a recent work solving a similar problem~\cite{Bang2021EstimatingData}.

\subsection{Metrics}

We evaluate the accuracy in predicting the number of panels within every pattern (\textbf{\#Panels}) and the number of edges within every panel (\textbf{\#Edges}). 
The cases in which the panel loop does not return to the origin are counted as having an incorrect number of edges as they usually require adding an edge to produce a connected shape. To estimate the quality of panel shape predictions, we use the average distance ($L2$ norm) between the vertices of predicted and ground truth panels with curvature coordinates converted to panel space and acting as panel vertices in this comparison (\textbf{Panel L2}). Similarly, we report $L2$ norm on the differences of predicted panel rotations (\textbf{Rot L2}) and translations (\textbf{Transl L2}) with ground truth values. The quality of predicted stitching information is described by a mean precision (\textbf{Precision}) and recall (\textbf{Recall}) of predicted stitches.   
\subsection{Comparing LSTM and Attention-based solutions for pattern shape recovery}
\label{sec:eval:generalization}

Here we compare the baseline hierarchical LSTM (\textbf{LSTM}) architecture, which relies on the global garment latent codes, versus the attention-based model for pattern shape recovery (\textbf{Att}). The presence of Stitch Tags in the output affects the performance of both models (Sec.~\ref{sec:eval:stitch_tag_ablation}), so for a cleaner comparison of shape recovery, we train and compare both architectures without stitch tags. We evaluate the models on the test set consisting of unseen garment examples of the same types that were used during training and on completely new types, according to the split described in Sec.~\ref{sec:data:split}. As can be seen from the results reported in Table~\ref{tab:shape_eval}, the baseline solution performs reasonably well on familiar types. However, it fails to generalize with less than 7\% success rate for predicting the correct number of panels in sewing patterns. An orderless loss (\textbf{LSTM Orderless}), which was found beneficial for RNN-based image multi-labeling task~\cite{OguzYazici2020OrderlessClassification}, did not improve the results. However, the attention-based solution showcases the ability to predict sewing patterns for the garment types unseen during training, correctly predicting the number of panels in more than 80\% of cases and performing well on other metrics. Examples of successful reconstructions are given in the supplementary materials, Fig.~\ref{fig:sup:test_unseen_results} and Fig.~\ref{fig:sup:test_seen_results}. In addition, the attention-based solution produces better panel shape quality (1.5 versus 2.7 Panel L2 for LSTM) on the test set but with somewhat less accurate panel placement.

\subsection{Choosing the method for stitch prediction}
\label{sec:stitches_eval}

To faithfully compare the two solutions for stitch prediction -- stitch tags as edge features and separate edge pairs classifier -- we evaluate both models on attention-based pattern shape prediction solution. We either train it jointly with stitch tags or use a trained pattern shape prediction model to produce inputs for the stitch model. The LSTM-based method does not generalize to novel garment types (Sec.~\ref{sec:eval:generalization}); hence it cannot provide suitable inputs for stitching information recovery methods for those cases.  In the case of unseen garment types, we evaluate precision and recall scores only on the outputs with a correctly predicted number of panels to prevent errors of the shape prediction model from affecting the stitch metrics.

The results of the evaluation are given in Table~\ref{tab:stitch-eval}. While stitch tags give near-perfect predictions on known types, their performance on unseen types is inferior to the best separate stitch model (\textbf{Model on Predictions}) in both precision and recall scores. In addition, training models with stitch tags seem to affect the performance of the shape prediction quality for both LSTM and attention-based solutions compared to the models trained without them (Table~\ref{tab:shape_eval}). Pattern shape model training with stitch tags is significantly slower (by about 15h on 2 GPU training). In contrast, the stitch classifier is lightweight, with 20 min training on a single GPU aided by two-hour inference of pattern shape model on the entire training set on a single GPU to obtain training data for stitch prediction.

We believe that the successful performance of a simple edge classifier model is explained by the fact that connected edges often have similar shapes, are close to each other in 3D, and stitches are often concentrated around certain body areas in our dataset. The latter might not hold for more complex garment types; hence a more sophisticated solution might be needed in the future. Another essential feature of the model is its reliance on only the local context of a potential stitch, which allows for a good performance on unseen garment types.

\subsubsection{Training set source for stitch prediction}
\label{sec:eval:stitch_train_set}

There are two options for choosing the type of input source to train connectivity prediction: using the edge vectors from ground truth panels or using the edges outputted by the trained pattern shape prediction module. We experimented with both approaches and then tested both models on the set of patterns outputted by the pattern shape prediction module as part of the integrated pipeline (Table~\ref{tab:stitch-eval}). As expected, the stitch model trained on the pattern shape predictions is more robust to the noisy inputs than the model trained on clean ground truth edges, hence having a significantly better performance for both seen and unseen garment types.

\subsection{Loss ablation study}

Most of the losses we use for training pattern shape prediction or stitching are indispensable components, without which the network will output random values for the corresponding variables. 
However, we find the need to justify the usage of loop loss for panel shape predictions and discuss encouraging segmentation-like behavior in attention scores and the effect of stitch tags prediction on shapes.

\subsubsection{Loop loss}

\begin{figure}[ht]
  \centering
  \includegraphics[width=0.8\linewidth]{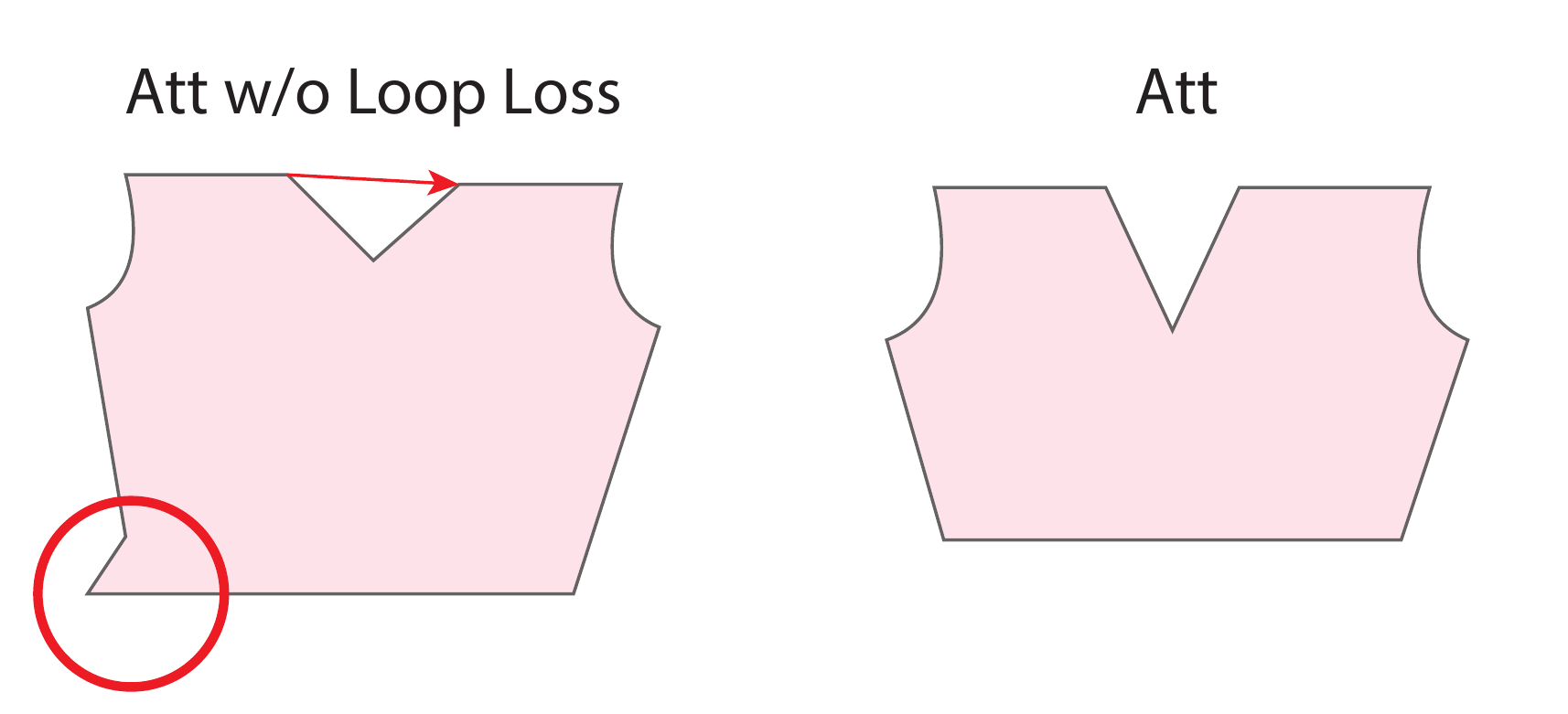}
  \caption{The front top panel of the dress sewing pattern as predicted by models trained with and without loop loss. The loop loss encourages the completion of shapes and reduces the relative shifting of edges.}
  \label{fig:eval:loop_loss}
\end{figure}

A model trained with only the basic losses, excluding the loop loss, tends to produce panels with the last edge not perfectly connected to the first one (Fig.~\ref{fig:eval:loop_loss}). Loop loss helps alleviate this issue. Quantitative analysis also reflects this effect: the overall edge accuracy score drops drastically when the loop loss is removed from the training process (Table~\ref{tab:shape_eval}).


\begin{figure*}[ht]
  \centering
  \includegraphics[width=\linewidth]{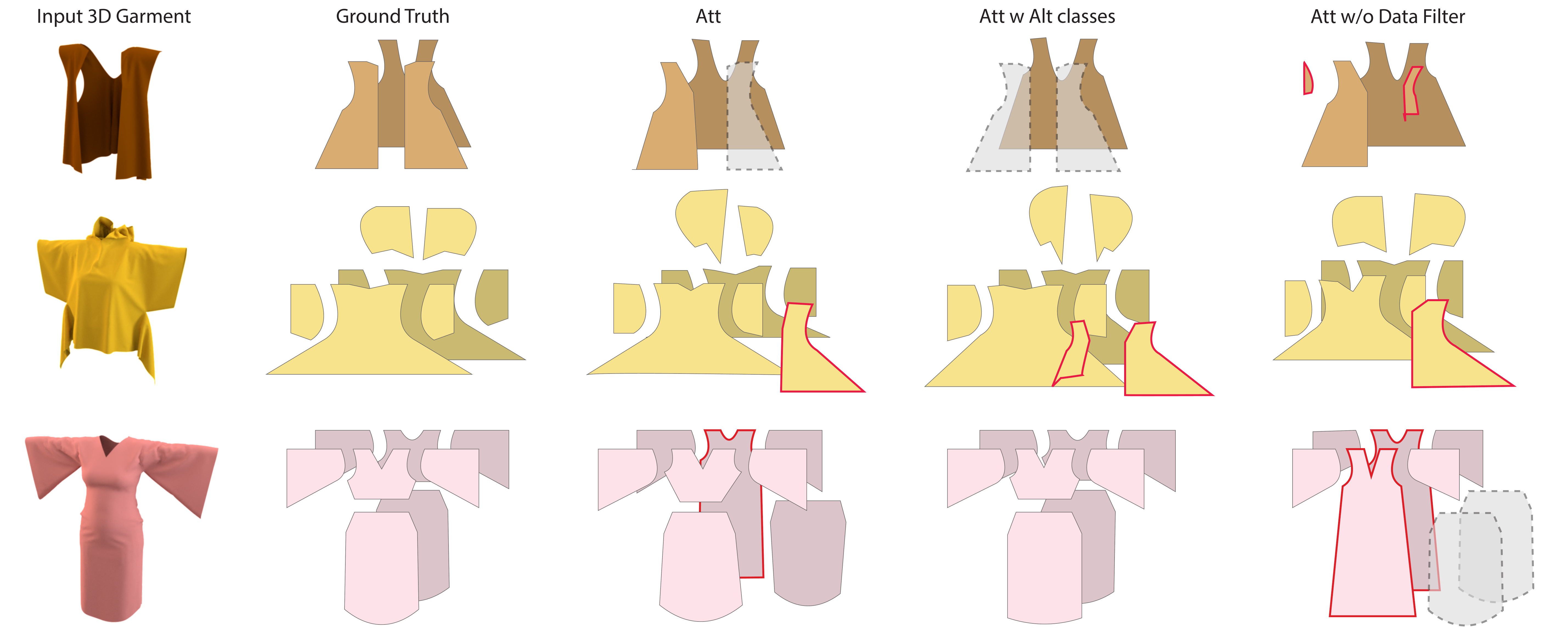}
  \caption{Examples of mispredicting sewing pattern topologies according to different data pre-processing strategies. Gray panels denote panels missing from prediction, and red outlines indicate extra panels or panels with significant errors.}
  \label{fig:topology_fails}
\end{figure*}  

\subsubsection{Effect of stitch tags presence}
\label{sec:eval:stitch_tag_ablation}

Experiments for both LSTM and Attention-based architectures show that the presence of stitch tags in the model negatively affects the quality of predicted panel shapes (Table~\ref{tab:shape_eval}). We conjecture that the reduction of complexity of both the output and the loss function when stitch tags are removed helps improve training for the rest of the sewing pattern features. These experiments provide an additional reason to prefer the separate stitching information regression model to the stitch tags-based solution.

\subsubsection{Learned attention scores vs. segmentation}

The way we formulate the attention scores (Sec.~\ref{sec:net:attention}) is very similar to the typical formulation for 3D model segmentation output layers. We also had an intuition that attention scores should resemble segmentation because they should attend to the area of the input 3D model corresponding to that particular panel location. With these two factors, it seems natural to encourage this segmentation-like behavior in attention scores explicitly using segmentation loss and ground truth segmentation labels provided with the dataset. 

We implement segmentation loss on SparseMax scores as introduced in the original paper~\cite{pmlr-v48-martins16} and add to the other losses on pattern shape with weighting to balance the scales of errors. The point cloud segmentation of inputs is transferred from the original mesh segmentation by taking the class of the nearest neighbor of each sampled point.

Results for the model trained on Segmentation loss (\textbf{Att w Segm}) are reported in Table~\ref{tab:shape_eval}. Unfortunately, segmentation loss was detrimental to performance on both parts of the test set. A closer inspection of attention scores generated by the model trained without segmentation loss (Fig.~\ref{fig:att_weights}) reveals that while attention tends to concentrate on the areas close to the corresponding panel, it also spreads to the areas of a panel's symmetric counterpart and to the areas that reveal related global features (e.g., the number of edges in top front or back panels depends on whether they connect to pant or skirt panels). Encouraging segmentation behavior may disturb the network from discovering these or similar dependencies.

\subsection{Effects of dataset preprocessing}

As was introduced in Sec.~\ref{sec:dataset}, we grouped the panels that compose the sewing patterns of garments in the dataset in classes and additionally filtered samples of the dataset to reduce the design overlap issues. Here we investigate the effect of these decisions.

\subsubsection{Panel classes}
\label{sec:eval:more_classes}

Panels are grouped in classes by their role in the garment and location around the body. For example, we grouped the front panels covering the trunk from t-shirts, dresses, and jumpsuits in one class (the full classification is given in the supplementary materials). However, there are several different ways to assign classes, and we found that the choice might affect the performance of the final panel shape prediction. To demonstrate this effect, we constructed an alternative set of classes where both panels that correspond to the left and right opening of the jackets are assigned a separate class. The original classification grouped one of the sides with full front panels of t-shirts, dresses, and jumpsuits. The results given in Table~\ref{tab:shape_eval} demonstrate that this extended class arrangement (\textbf{Att w Alt Classes}) has worse performance on unseen garment types. In Fig.~\ref{fig:topology_fails} we demonstrate failure cases for these experiments to showcase qualitative differences. For example, when the original classification misses or adds one side of the jacket, the alternative ones miss or add both. 

These observations lead to the conclusion that the question of panel classification might be more complex than we initially thought and thus need to be investigated further in more detail in future work.

\subsubsection{Sample filtering}

Table~\ref{tab:shape_eval} shows the results of the attention-based pattern shape prediction as trained on the full dataset without sample filtering (\textbf{Att w/o Data Filter}). The performance of this model on the filtered test set is comparable to the results of our main model (Att), although the panel shape metric is slightly worse on both seen and unseen types. However, we should note that removing filtering, increased the number of garment samples available for training by about two times, which affects the performance as well.

It is noteworthy that the model trained on the full dataset produces different quality errors compared to the original run. Fig.~\ref{fig:topology_fails} shows a prediction of extra sleeve panels for sleeveless but wide garment examples or misinterpretation of a dress as a long t-shirt pattern. These mistakes 
appear to be the issues of design overlap, which we mentioned in Sec.~\ref{sec:data:filter}. 

\subsection{Comparison with flattening-based sewing pattern recovery}


\begin{figure*}[ht]
  \centering
  \includegraphics[width=\linewidth]{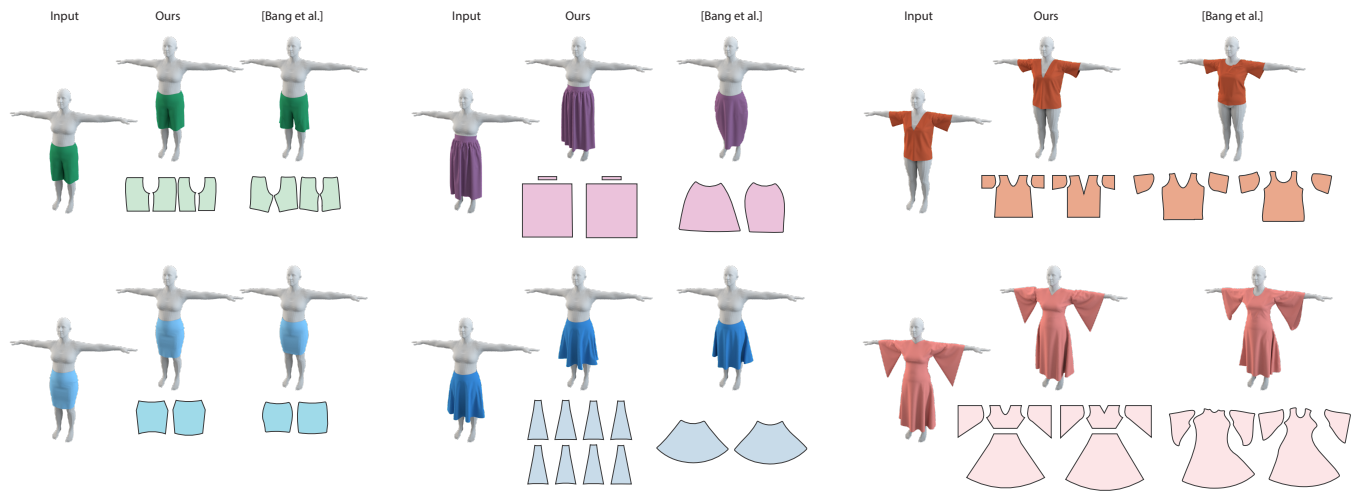}
  \caption{Comparison of sewing pattern reconstruction of NeuralTailor and the method of Bang et al.~\shortcite{Bang2021EstimatingData}. Our method produces patterns with cleaner shapes, shows the capability to handle complex panel arrangements (second column), and reproduces garment fit (third column). Best viewed on screen zoomed-in.}
  \label{fig:evl:compare_with_flattening}
\end{figure*}  

The work in the area of sewing pattern shape recovery is somewhat limited, especially when focusing on the direct estimation of sewing patterns from geometry without access to initial guesses or templates. Our closest competitor is the work of Bang et al.~\shortcite{Bang2021EstimatingData} that uses a flattening-based approach and shows generalization across garment types.

We compare this work with the predictions of NeuralTailor on garments from our test set. Note that our examples present favorable conditions to the work as they contain clean and full geometry with minimum distortions due to human poses. We chose garment samples from types supported by the original work (no jackets, hoods, or jumpsuits). Figure~\ref{fig:evl:compare_with_flattening} demonstrates that while both methods give good results on simple garments (pants, pencil skirt), our method consistently estimates the garment fit better (pants, t-shirt, dress), while the competing technique produces slightly looser garments. NeuralTailor can also handle stylistic panel arrangements that are difficult to recover by the body-part-based cuts (skirt with a belt and flared skirt). Sewing patterns produced by our method are symmetric and contain straight lines and simpler curves, reproducing typically expected sewing pattern shapes.

\subsection{Robustness to input noise}
\label{sec:eval:robustness}

To get a complete picture of the NeuralTailor behavior, we test its robustness to different types of noise present in the input.

\subsubsection{3D Scanning Artifacts Imitation}

Alongside the clean meshes, the dataset contains their corrupted versions. The corruption imitates artifacts of 3D scanning -- missing geometry in the areas invisible to the capturing camera, which is a typical problem for heavy folds (e.g., Fig. 2 in~\cite{Korosteleva2021GeneratingPatterns}). We found that of NeuralTailor is robust to this type of noise (Table~\ref{tab:shape_eval}) despite being trained on full geometry only. Most likely, the randomness and sparsity of point clouds sampled for training encouraged the model to learn how to handle missing geometry. Another reason could be the attention mechanism itself. It focuses the processing on an even smaller number of points, which increases the chances of avoiding the areas where gaps are likely to occur.

\subsubsection{Gaussian Noise}

\begin{figure*}[ht]
  \centering
  \includegraphics[width=0.95\linewidth]{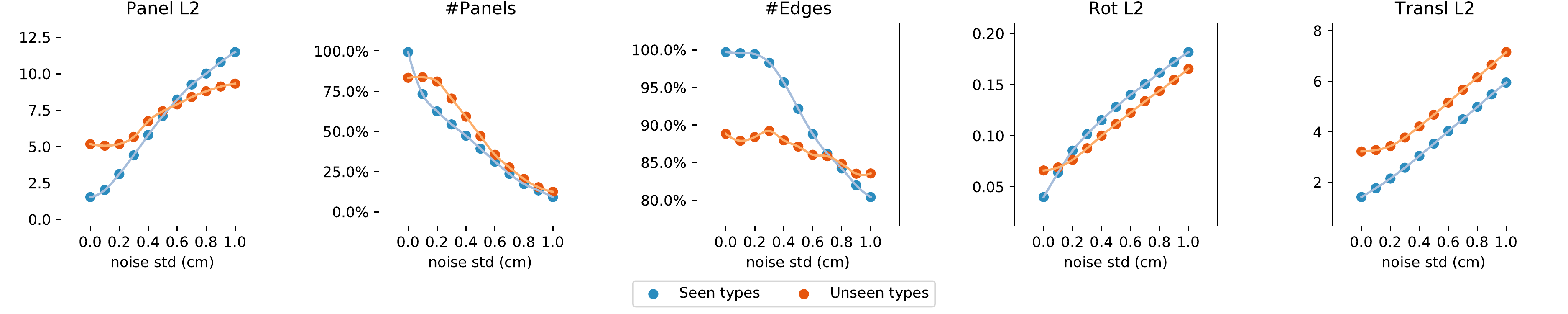}
  \caption{NeuralTailor pattern shape reconstruction performance on seen (blue) and unseen (orange) garment types depending on the noise level in the input point cloud. The noise is given as Gaussian noise with a specified standard deviation.}
  \label{fig:noise_lev}
\end{figure*}

In our synthetic point clouds, all the sampled points are located precisely on the surface of the corresponding 3D models. On the other hand, it is natural to expect noise in the points' locations for the in-the-wild 3D scans. We approximate such artifacts by adding Gaussian noise to the point locations of the point clouds in the test set, varying the standard deviation between 0 and 1 cm. Evaluation of NeuralTailor on these inputs (Figure~\ref{fig:noise_lev}) demonstrates performance drop as the noise gets more severe. Hence, NeuralTailor is only capable of handling small levels of noise, as present in high-precision 3D scanning systems. 
Improving the robustness to handle noisier data, e.g., from systems like Kinect or single-view RGB reconstructions, would be an interesting direction for future work.

\subsubsection{In-the-wild data}
\begin{figure}[t]
  \centering
  \includegraphics[width=0.8\linewidth]{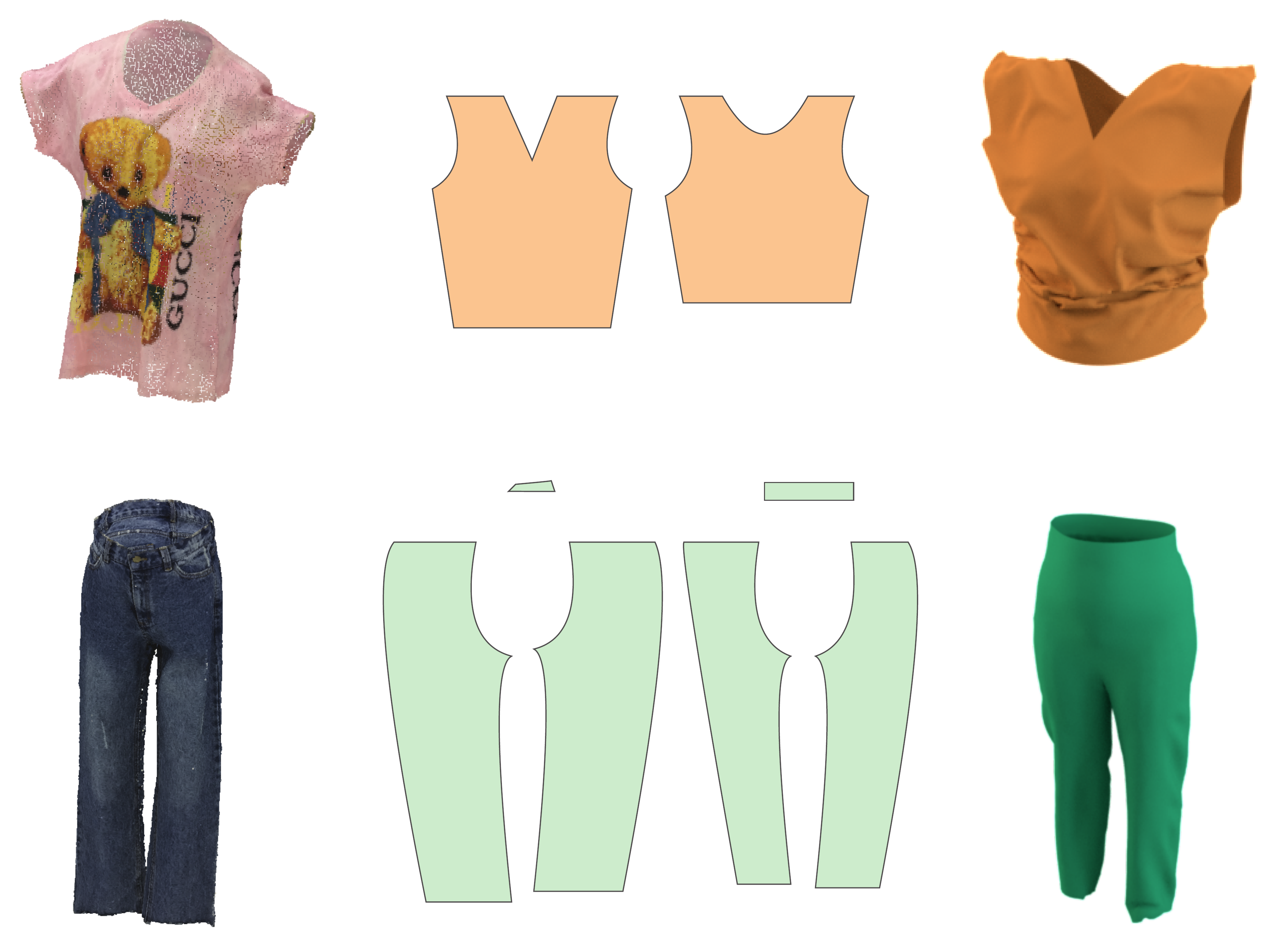}
  \caption{Sewing patterns predicted by NeuralTailor on examples from Deep Fashion3D dataset.
  For jeans, the panels for the belt (two panels on top) were refined to proper shapes before draping. 
  }
  \label{fig:real_data_cut}
\end{figure}

We qualitatively evaluate the framework on the garment captures from Deep Fashion3D dataset~\cite{Zhu2020} (Fig.~\ref{fig:real_data_cut}). We found that the model makes relatively good guesses about the garment structure; for example, jeans got a pattern of classic pants with a belt, a type that was not included in our dataset. On the other hand, the quality of pattern shape and panel placement prediction degraded on the in-the-wild scans, and bridging this sim-to-real gap will be an important direction of future work.


\section{Discussion and Future Work}
\label{sec:discusstion}

This paper presented NeuralTailor, the first learning-based solution for recovering sewing patterns as structures from 3D garment point clouds. We introduced a baseline based on hierarchical LSTMs capable of recovering sewing patterns for drastically different garment types within the same model. We then suggested a novel attention mechanism and a stitch recovery module that both focus on exploring local features to enable generalization to novel garment types and sewing pattern topologies unseen during training.

This work is the first step toward neural sewing pattern recovery. It successfully demonstrates how structured representation and consideration for the local context could allow generalization beyond the data available for training, which is particularly useful for the ever-evolving garment domain. 

However, there are several directions left for future research. There is a need for additional solutions to handle overlapping designs and, for example, make the framework produce multiple or any of the plausible patterns for a particular input garment. It would also be interesting to explore if the optimal panel clustering can be found automatically instead of relying on a heuristic decision. The current stitching model prediction, although successful, might benefit from additional exploration and direct access to the input 3D point cloud. 

Another consideration for future work is an exploration of symmetrical properties of garments, such as left-right symmetry or symmetry of edges in the stitches. Since symmetry could be violated for stylistic purposes, we avoided incorporating it into our system for generalizability. However, the simplicity of our output sewing pattern structures allows enforcing symmetry in post-processing, e.g., finding corresponding edges by relying on their 3D positions and matching their lengths. Such a post-processing step may further improve the quality of predicted sewing patterns.

On a higher level, an important direction would be to bring the pipeline closer to in-the-wild data. NeuralTailor would improve its applicability by building stronger resistance to noise in the input point clouds, incorporating variations due to material properties, body poses, or shapes, and fine features of sewing pattern design, such as darts, pleats, or complex edge curves. On top of that, it would be important to consider complex garment arrangements, such as heavy occlusions due to garment layering, layering of fabric (e.g., ballroom skirts), accessories that change the standard draped shape like belts, and utilization of complex materials like thick and bumpy winter coats. Bringing these features would require an extension of the currently available datasets. Other technologies, such as domain transfer and adaptation, could also be worth exploring to reduce the need for data labeled with ground truth sewing patterns.

\begin{acks}

 We are deeply grateful to the members of Geomteam of LAVA Lab, KAIST, for their continued support and invaluable discussions throughout this journey. Many thanks to Seungbae Bang for providing the results of his flattening-based approach on our garment data. Finally, we would like to thank the anonymous reviewers for their valuable time and suggestions. This work was supported by ITP, MSIT, Korea (2022-0-00566) and NST, MSIT, Korea (CRC 21011).

\end{acks}   

\bibliographystyle{ACM-Reference-Format}
\bibliography{references/additional.bib,references/mend_copy.bib}  

\clearpage
\appendix

\section*{NeuralTailor: Appendix}

\begin{figure*}[b]
  \centering
  \includegraphics[width=0.8\linewidth]{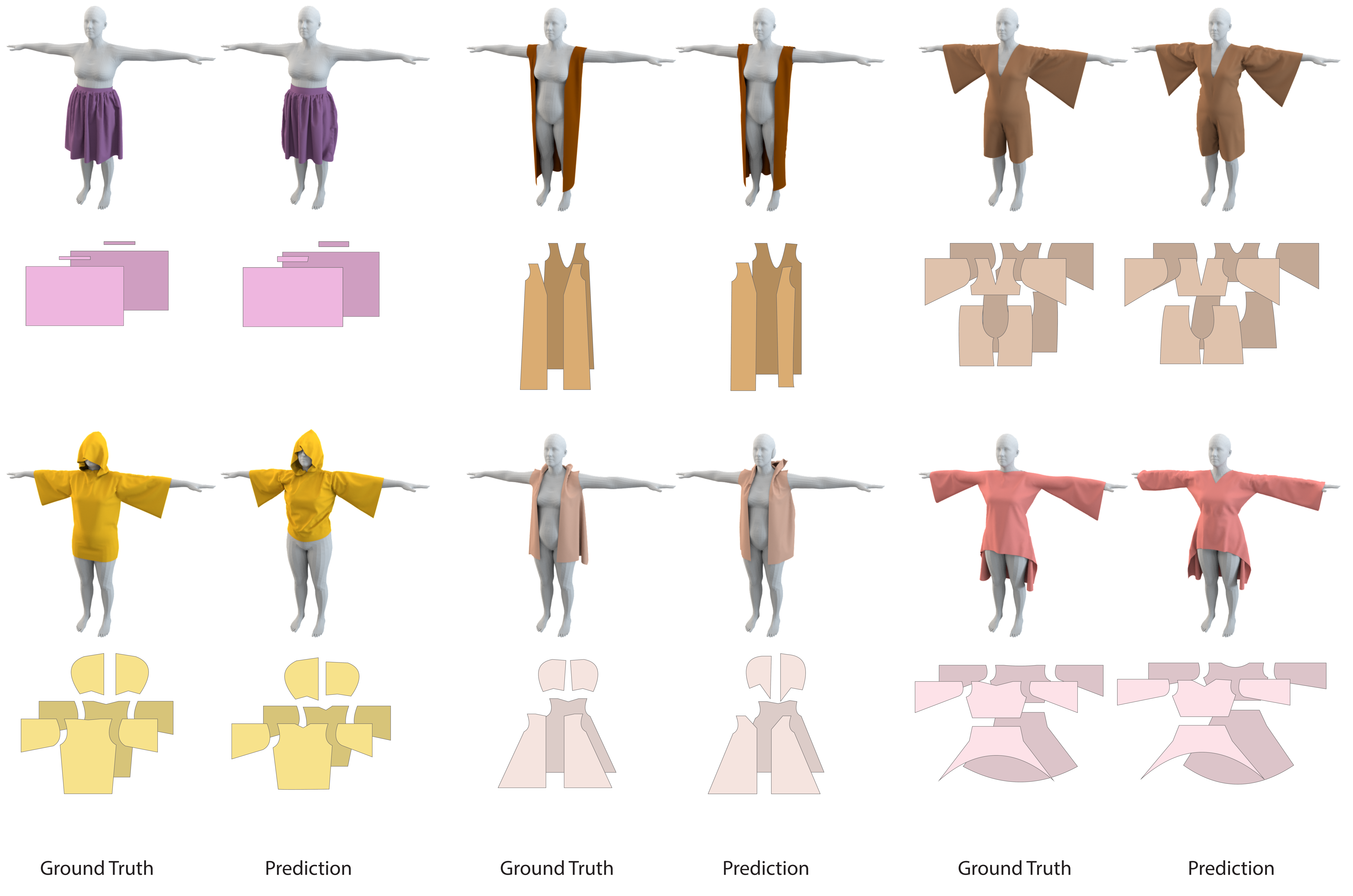}
  \caption{Example sewing pattern predictions by NeuralTailor on garment types unseen during training. Stitches are not shown for clarity of visualization. Predicted sewing patterns are draped on SMPL~\cite{Loper2015} woman body model using a stand-alone physics simulator.}
  \label{fig:sup:test_unseen_results}
\end{figure*}  

\section{Implementation details}
\label{sec:supp:implementation}

\subsubsection*{Architecture details}
Each EdgeConv layer in both of our architectures (Figures~\ref{fig:net:baseline},~\ref{fig:net:neural_tailor_full}) uses a small MPL with two hidden layers of 200 neurons each and an output layer of 150 neurons. The dynamic graphs on each layer are constructed using $k=5$ nearest neighbors. Per-point features are aggregated from edge features using max-pooling. The final per-point feature has the size of 153 thanks to skip connection with input point coordinates. Pattern LSTM (Sec.~\ref{sec:baseline:rnn_enc}) cell contains two layers with 250 elements per hidden layer and output of the same size. Attention MLP (Sec.~\ref{sec:net:attention}) consists of 3 layers with 153 neurons each and outputs vector of size 31 (number of panel classes) or 32 for the experiment with additional class (Sec.~\ref{sec:eval:more_classes}). The panel decoder's LSTM cell contains three layers with 250 elements per layer, outputting edge features of size 4 or size 8 if stitching information is included. The maximum number of edges generated by PanelLSTM is 14. The MLP for decoding the panel placement consists of one linear layer mapping the 250-element panel encoding to the vector of 7 elements representing concatenated rotation and translation. The stitch information prediction model (Sec.~\ref{sec:net:stitch_model}) is MLP with 3 layers, a hidden layer size of 200 and an output layer of one neuron. For training this model, we sample 200 edge pairs that are stitched and 200 that are not from each sewing pattern example in the batch. All LSTM cells operate in a one-to-many manner. We construct the input to LSTM as a sequence of duplicated input encodings for the desired length of the output sequence. 

\subsubsection*{Data pre-processing}
The 3D garment models from the dataset of~\cite{Korosteleva2021GeneratingPatterns} have a clean mesh structure with visible seam lines. We randomly sample point clouds from the surface of these models. Each sample point cloud contains 2000 points. To stabilize training, we additionally apply standardization (bringing mean to zero and standard deviation to one) on input point clouds, edge vectors, curvature coordinates, and normalization (ensuring all values are between 0 and 1) on panel rotations and translations. 

\subsubsection*{Training settings} We found it beneficial to use one-cyclic learning rate scheduling, following recommendations of~\cite{Smith2018ADECAY}, with the maximum learning rate of 0.002. We train all models for 350 epochs with Adam optimizer~\cite{Kingma2015} and batch size of~30 with early stopping enabled for when the model does not improve for consecutive 100 epochs. The training pipeline is implemented with PyTorch~\cite{PyTorchNEURIPS2019_9015}, PyG~\cite{PyG/Fey/Lenssen/2019}, and Weights and Biases~\cite{wandb}.

\subsubsection*{Training times}
Training the pattern recovery model without stitch tags takes about 36 hours on two NVIDIA Titan Xp GPUs. Training of the stitch model follows a similar training setup and takes about 2 hours for inference of pattern shapes from the pattern recovery model on the training set and 30 min of the actual training on a single NVIDIA Titan XP GPU. The full baseline model with LSTM backbone and stitch tag recovery takes 72 hours to train on two NVIDIA Titan Xp GPUs.

\section{Example predictions on the test set}

\begin{figure*}[htb]
  \centering
  \includegraphics[width=0.9\linewidth]{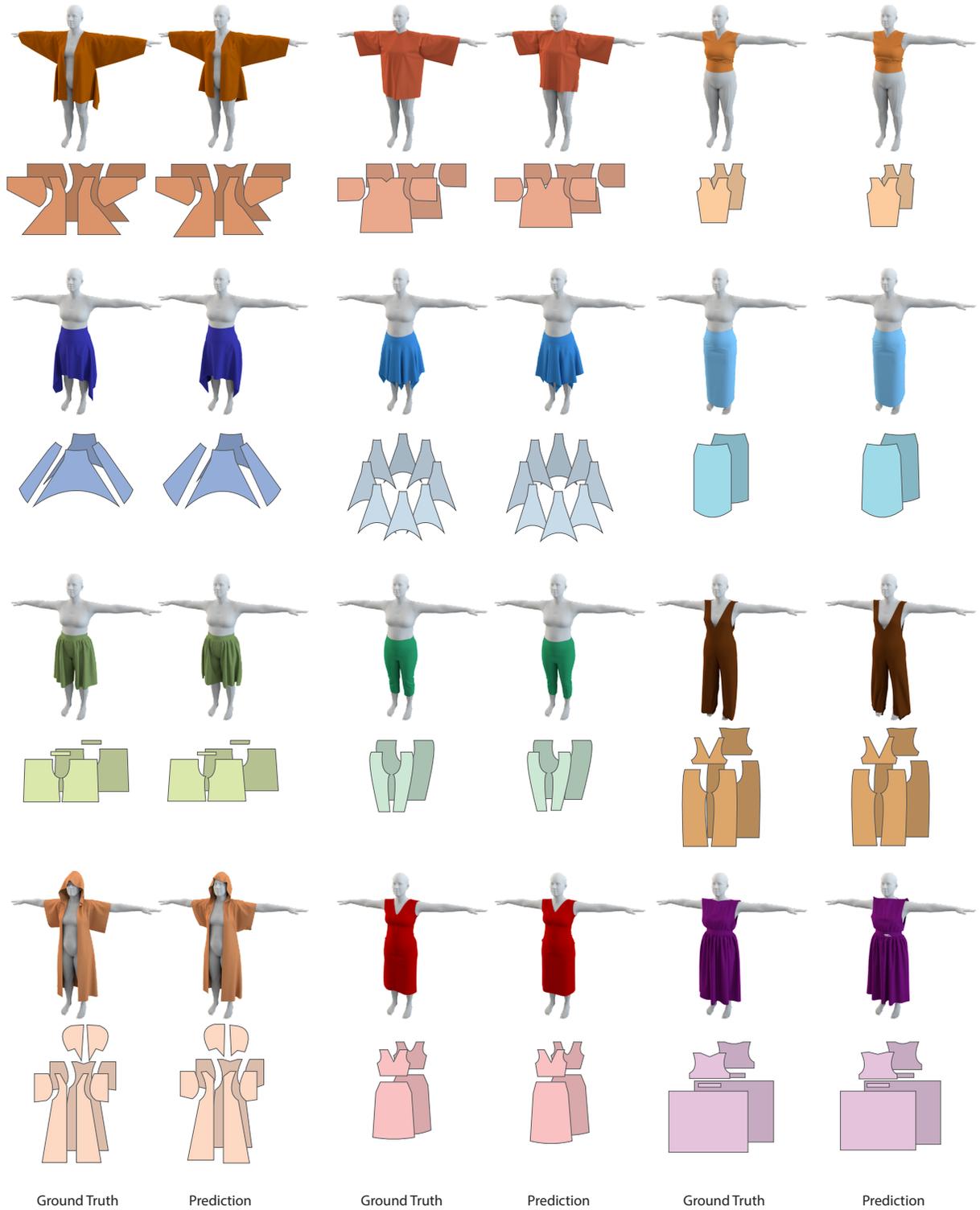}
  \caption{Example sewing pattern predictions by NeuralTailor for garment types available during training, one example of each type. Stitches are not shown for clarity of visualization. Predicted sewing patterns are draped on SMPL~\cite{Loper2015} woman body model using a stand-alone physics simulator.}
  \label{fig:sup:test_seen_results}
\end{figure*}


\end{document}